\documentclass[sn-mathphys,Numbered]{sn-jnl}

\usepackage{graphicx}%
\usepackage{multirow}%
\usepackage{amsmath,amssymb,amsfonts}%
\usepackage{amsthm}%
\usepackage{mathrsfs}%
\usepackage{xcolor}%
\usepackage{textcomp}%
\usepackage{manyfoot}%
\usepackage{multirow}
\usepackage{algorithm}%
\usepackage{algorithmicx}%
\usepackage{algpseudocode}%
\usepackage{listings}%
\usepackage{caption}
\usepackage{booktabs}

\newcommand{\ie}{\emph{i.e.}}
\newcommand{\eg}{\emph{e.g.}}
\newcommand{\etal}{\emph{et al.}}

\begin{document}

\title[Article Title]{Cascaded Cross-Modal Transformer for Audio-Textual Classification}


\author[1]{\fnm{Nicolae-C\u{a}t\u{a}lin} \sur{Ristea}}

\author[1]{\fnm{Andrei} \sur{Anghel}}

\author[2]{\fnm{Radu Tudor} \sur{Ionescu}}

{\equalcont{Corresponding author: {raducu.ionescu@gmail.com}.}}


\affil[1]{\orgdiv{National University of Science and Technology Politehnica Bucharest}, \orgaddress{\country{Romania}}}

\affil[2]{\orgdiv{University of Bucharest}, \orgaddress{\country{Romania}}}


\abstract{
Speech classification tasks often require powerful language understanding models to grasp useful features, which becomes problematic when limited training data is available. To attain superior classification performance, we propose to harness the inherent value of multimodal representations by transcribing speech using automatic speech recognition (ASR) models and translating the transcripts into different languages via pretrained translation models. We thus obtain an audio-textual (multimodal) representation for each data sample. Subsequently, we combine language-specific Bidirectional Encoder Representations from Transformers (BERT) with Wav2Vec2.0 audio features via a novel cascaded cross-modal transformer (CCMT). Our model is based on two cascaded transformer blocks. The first one combines text-specific features from distinct languages, while the second one combines acoustic features with multilingual features previously learned by the first transformer block.
We employed our system in the Requests Sub-Challenge of the ACM Multimedia 2023 Computational Paralinguistics Challenge. CCMT was declared the winning solution, obtaining an unweighted average recall (UAR) of $65.41\%$ and $85.87\%$ for complaint and request detection, respectively. Moreover, we applied our framework on the Speech Commands v2 and HarperValleyBank dialog data sets, surpassing previous studies reporting results on these benchmarks. Our code is freely available for download at: \url{https://github.com/ristea/ccmt}.
}




\maketitle

\section{Introduction}\label{sec1}
In recent years, the field of audio classification has witnessed significant advancements in analyzing and interpreting verbal and non-verbal vocal cues, leading to valuable insights into human communication. However, with limited training data, it is often hard to learn robust features solely from the audio domain. In the current research landscape, we introduce a multimodal framework for speech and text processing. Notably, our method constructs multiple text representations (modalities) from audio data using pretrained models, essentially requiring only the audio modality as the original input. Since the text modalities are generated by neural models, the proposed method is well-suited for a wide range of speech classification tasks, ranging from complaint detection to intent recognition. 

\begin{figure*}[!t]
\begin{center}
\centerline{\includegraphics[width=1.0\linewidth]{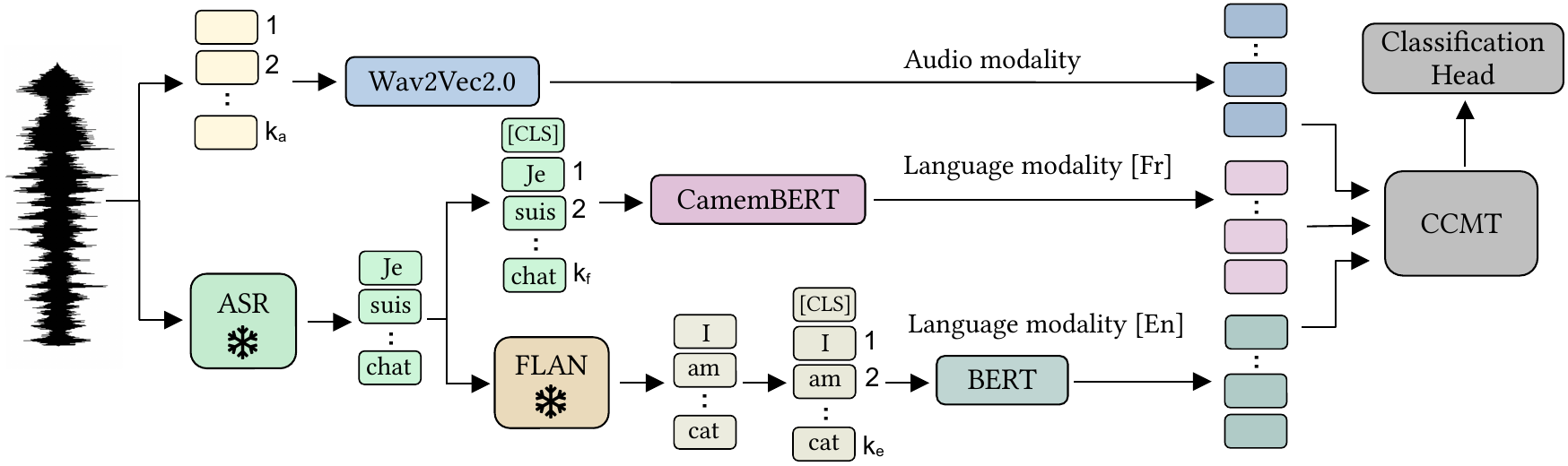}}
\vspace{0.2cm}
\caption{Our multimodal pipeline for speech classification. Audio is the original input modality, from which we extract tokens with the Wav2Vec2.0 \cite{Baevski-NeurIPS-2020} network that processes audio data in the time domain. To generate the first extra (text) modality, we initially employ an ASR model to transcribe each audio sample into text. We assume that the spoken language is known a priori, and, in the depicted example, the audio is in French [Fr]. The French text is given as input to the CamemBERT \cite{Martin-ACL-2020} language model. The next step is to generate the second text modality, which relies on the FLAN \cite{Chung-ARXIV-2022} model to translate the French transcripts into English. For the English language modality [En], the text is processed by the BERT model \cite{Devlin-NAACL-2019}. The audio tokens returned by Wav2Vec2.0 and the text tokens produced by CamemBERT and BERT are further fed into our cascaded cross-modal transformer (CCMT). The final class token provided by CCMT is fed into the classification head. Our framework operates similarly if the original spoken language is English. The frozen models are marked with a snowflake.}
\label{fig_framework}
\end{center}
\end{figure*}

Taking inspiration from the success of existing multimodal methodologies \cite{Akbari-NeurIPS-2021, Das-ACM-2023, Georgescu-arXiv-2022, Jabeen-TACM-2023, Yoon-SLT-2018} in other domains, we propose a novel audio-textual pipeline which effectively harnesses multimodal features derived from both speech and text data, which are further integrated into a cascaded cross-modal transformer (CCMT), as shown in Figure \ref{fig_framework}. To generate multimodal representations from the audio modality, \ie~the only input modality considered in our work, we employ state-of-the-art automatic speech recognition (ASR) models \cite{Baevski-NeurIPS-2020, Radford-ARXIV-2022} to transcribe the audio samples. The additional text modality, obtained through speech-to-text conversion, can provide valuable insights that complement the original audio data, enabling the use of a wide range of natural language processing (NLP) methods. Furthermore, recognizing the existence of large language models (LLMs) tailored for distinct languages \cite{Martin-ACL-2020, Devlin-NAACL-2019, Canete-ICLR-2020}, we enlarge the scope of our framework by translating the transcripts into multiple languages, such as English, French and Spanish, via neural machine translation (NMT). To accomplish this, we utilize the FLAN T5 language model \cite{Chung-ARXIV-2022}, which enables us to effectively analyze the same data in different linguistic contexts.

Taming the complexity of real-world classification tasks through the combination of various modalities is not an easy challenge, requiring the development of robust and efficient methods to effectively fuse multiple data sources \cite{Ramachandram-SPM-2017, Gao-NC-2020, Stahlschmidt-BB-2022}. To address this challenge, we propose the CCMT model comprising successive transform blocks which aggregate modalities two by two. More specifically, we consider two text modalities (English and French) and one audio modality, which are combined via two cascaded transform blocks. The first block aggregates information from two LLMs, namely BERT \cite{Devlin-NAACL-2019} and CamemBERT \cite{Martin-ACL-2020}. The second block combines the resulting multilingual text features with audio-based Wav2Vec2.0 \cite{Baevski-NeurIPS-2020} features. By integrating audio-textual information, our model captures the rich semantic and acoustic characteristics of the data, enabling a comprehensive understanding of the original audio data. On the one hand, the employed NLP models facilitate capturing nuanced language cues and contextual features within conversations. On the other hand, the audio model can extract information about vocal tone, emphasis, and other non-verbal cues, providing insights about complementary features that contribute to the overall assessment of the analyzed speech.

Our preliminary study presented at ACMMM 2023 \cite{Ristea-ACMMM-2023} provided a brief description of CCMT and its results in the Requests Sub-Challenge of the ACM Multimedia 2023 Computational Paralinguistics Challenge \cite{Schuller-ACMMM-2023}. Remarkably, our model is the winner of the Requests Sub-Challenge\footnote{\url{http://www.compare.openaudio.eu/winners/}}, surpassing the second-best model \cite{Porjazovski-ACMMM-2023} by $3\%$. We hereby extend our preliminary work \cite{Ristea-ACMMM-2023} with a more comprehensive presentation of the method, as well as a more extensive evaluation, considering additional benchmarks and ablation studies. Indeed, we conduct experiments on Speech Commands v2 and HarperValleyBank to show that CCMT is beneficial for a broad range of speech classification tasks. Our empirical results show that CCMT outperforms existing methods on all benchmarks.   

In summary, our contribution is threefold:
\begin{itemize}
    \item We propose a novel speech classification pipeline that employs ASR and NMT to generate multiple text modalities from speech samples, which empowers the neural network to harness multiple linguistic representations.
    \item We introduce a novel audio-textual model which combines audio and text representations via a cascaded cross-modal transformer, termed CCMT.
    \item We conduct comprehensive experiments on three distinct data sets, obtaining strong empirical results that support our new design. 
\end{itemize}

\section{Related Work}

As related work, we discuss the recent studies treating the audio and text modalities in a joint or independent manner.

\subsection{Audio classification}
In recent years, deep learning has emerged as a prominent approach in the audio domain, thanks to the advancements in deep neural network architectures \cite{Purwins-JSTAR-2019, Ristea-INTERSPEECH-2020, Kong-TASLP-2020, Gong-INTERSPEECH-2021, Ristea-INTERSPEECH-2022} and the availability of large-scale audio data sets \cite{Gemmeke-ICASSP-2017}. A common approach involves transforming audio examples into image-like representations, such as spectrograms obtained via the short-time Fourier transform (STFT), followed by the application of a convolutional neural network (CNN) for the desired task. The ability of CNNs to capture local spectral and temporal features makes them well-suited for analyzing audio signals. For instance,  in \cite{Ristea-INTERSPEECH-2020}, the authors proposed a CNN-based ensemble using a support vector machines (SVM) meta-model applied on the learned embedding space. This solution achieved promising results in a previous ComParE challenge, surpassing the baseline proposed by the organizers by $2.8\%$. Alongside CNN architectures, attention mechanisms have garnered increasing interest within the audio domain \cite{Gong-INTERSPEECH-2021, Ristea-INTERSPEECH-2022, Gong-AAAI-2022, Huang-NeurIPS-2022}. Attention mechanisms enable models to focus on informative parts of the input and capture both local and global contextual information. Gong \etal~\cite{Gong-INTERSPEECH-2021} adapted the vision transformer model \cite{Dosovitskiy-ICLR-2020} to the audio domain by dividing the input spectrogram into overlapping patches. Their approach outperformed previous methods, while employing a convolution-free architecture. Similarly, Ristea \etal~\cite{Ristea-INTERSPEECH-2022} proposed a separable transformer architecture, treating the time and frequency bins as separate input tokens. Our work integrates a hybrid architecture known as Wav2Vec2.0 \cite{Baevski-NeurIPS-2020}, which is based on both convolutional and transformer blocks. It processes audio data in the time domain, extracting features from distinct overlapping windows with a CNN-based model, and treating the resulting embeddings as input tokens for a transformer. Unlike existing audio classification models, we integrate multiple pretrained models to generate additional modalities and benefit from the prior knowledge learned by audio and text processing models.


\subsection{Text classification}
Text classification is a fundamental task in the field of NLP, gaining steadily increasing attention, particularly with the emergence of transformer models. Over the years, numerous approaches have been proposed to tackle this generic task, employing a variety of techniques and models \cite{Devlin-NAACL-2019}. Transformer-based models \cite{Vaswani-NIPS-2017}, notably the Bidirectional Encoder Representations from Transformers (BERT) \cite{Devlin-NAACL-2019}, have revolutionized text classification by leveraging self-attention mechanisms in capturing contextual information from both preceding and subsequent words. BERT has achieved state-of-the-art performance across various NLP tasks \cite{Minaee-CSUR-2021, Gasparetto-I-2022} and has become widely adopted in text classification. Furthermore, researchers have explored techniques such as incorporating domain-specific knowledge \cite{Khadhraoui-MDPI-2022, Wan-JS-2022, Yang-ICASSP-2022} or training language-specific models \cite{Devlin-NAACL-2019, Martin-ACL-2020, Canete-ICLR-2020, Dumitrescu-EMNLP-2020} to enhance performance in downstream text classification tasks. These efforts have contributed to advancements in the field and improved the effectiveness of text classification approaches across various languages. 

    
With the advancements in both NLP and ASR fields, we consider that extending audio classification to audio-textual classification by transcribing audio files and applying NLP models, resulting in a multimodal framework, can lead to better results. However, there is a limited number of studies that harness the extracted text information to complement the audio modality \cite{Bhaskar-PCS-2015, Yoon-TOG-2020, Porjazovski-ACMMM-2023}. Bhaskar \etal~\cite{Bhaskar-PCS-2015} combined shallow and knowledge-based models for emotion recognition, while Yoon \etal~\cite{Yoon-TOG-2020} addressed the task of speech gesture generation, combining video, speech and text modalities. Closer to our work, Porjazovski \etal~\cite{Porjazovski-ACMMM-2023} extracted linguistic information and combined it with acoustic features via a late fusion model. Different from these related studies, we transcribe the audio files using multiple ASR models \cite{Radford-ARXIV-2022} to make our model robust to specific transcription errors. We further employ a model for text classification \cite{Chung-ARXIV-2022} (in the native language of the audio data), but we also translate the transcripts into different languages, \eg~English, French and Spanish. 
Our approach harnesses enriched representations and contextual understanding captured by multiple language-specific transformers, enhancing performance by aggregating and analyzing multimodal information. 

\subsection{Multimodal classification}
Multimodal deep learning has attracted significant attention in recent years as it enables effective modeling and analysis of complex data from multiple modalities. While research has extensively explored the fusion of audio-visual data \cite{Abdu-IF-2021, Georgescu-arXiv-2022, Pandeya-SENSORS-2021,Ramachandram-SPM-2017} and audio-visual-textual data \cite{Sun-AAAI-2020, Akbari-NeurIPS-2021}, with notable advancements in fusion techniques, the combination of audio and text has received comparatively less attention \cite{Yoon-SLT-2018, Singh-KBS-2021, Toto-CIKM-2021, Porjazovski-ACMMM-2023}. Singh \etal~\cite{Singh-KBS-2021} proposed a hierarchical approach for multimodal speech emotion recognition, combining 33 audio features with textual features extracted from large language models. In \cite{Toto-CIKM-2021}, the authors introduced AudioBERT, a model that integrates pretrained audio and text representation models along with a dual self-attention mechanism. In contrast to existing methods, which only aggregate information from fine-tuned audio and text models, we propose to construct additional text modalities through automatic speech recognition and neural machine translation. 
By incorporating multilingual text branches, our model can capture nuances present in different languages, providing a more comprehensive representation of the original audio data.

One key aspect in multimodal learning is the fusion methodology. A wide variety of frameworks have been proposed in the literature, including approaches based on early, intermediate, and late fusion \cite{Boulahia-MVA-2021, Huang-NPJ-2020, Wang-NeurIPS-2020, Huang-ICASSP-2020, Li-CVPR-2023, Pawlowski-SENSORS-2023, Xu-TPAMI-2023, singh2021multimodal}. For instance, Li \etal~\cite{Li-CVPR-2023} introduced an efficient and flexible multimodal fusion method specifically designed for fusing unimodal pretrained transformers. Shvetsova \etal~\cite{Shvetsova-CVPR-2022} proposed a 
modality-agnostic fusion transformer that learns to exchange information among multiple modalities, such as video, audio, and text, and integrates them into a fused representation in a joint multimodal embedding space. In a different study, Lee \etal~\cite{Lee-ECCV-2022} presented a cross-modality attention transformer and a multimodal fusion transformer, which identify correlations between data modalities and perform feature fusion. In a similar fashion, Liu \etal~\cite{Liu-PRL-2023} proposed a cross-scale cascaded multimodal fusion transformer to facilitate interaction and fusion among modalities with multi-scale features. 
Singh \etal~\cite{singh2021multimodal} proposed a multimodal fusion framework for emotion recognition by using audio and text features captured by transcribing raw audio files. Similarly, Braunschweiler \etal~\cite{braunschweiler2022factors} proposed a multimodal emotion recognition system based on speech and transcribed text, which attained superior results compared to prior studies. Moreover, they showed that the multimodal audio-text fusion leads to better results. In a similar fashion, Sharma \etal~\cite{sharma2023real} proposed a multimodal framework based on video, audio and transcribed text for real-time emotional health detection. In contrast to the aforementioned works, our approach introduces a cascaded cross-attention transformer that incorporates attention across distinct languages, followed by cross-attention between the resulting linguistic features and the speech features. This design allows us to benefit from linguistic representations in different languages and effectively integrate them with the audio modality.

\section{Method}

We introduce a novel multimodal pipeline for speech classification, which is depicted in Figure \ref{fig_framework}. Given the audio input data, our pipeline produces two additional text modalities through automatic speech recognition (ASR) and neural machine translation (NMT). 
The proposed cascaded cross-modal transformer (CCMT) further processes the three modalities. Next, we provide a detailed explanation of our pipeline, separately addressing each component.

\subsection{Audio branch}

We utilize the pretrained Wav2Vec2.0 model \cite{Baevski-NeurIPS-2020} to learn representative tokens for the audio modality. The raw audio data is partitioned into $k_a \in \mathbb{N^+}$ chunks, where $k_a$ depends on the input length and differs from one sample to another. The model processes the initial tokens, leading to a meaningful audio representation. This process involves several stages, including preliminary convolutional layers for local feature extraction and transformer layers for global contextual information extraction.
The output of the Wav2Vec2.0 neural network has the same number of tokens (as the input), denoted as $k_a$, which represent the acoustic features of the audio modality. The resulting tokens encapsulate features about the audio signals, including pitch, frequency, and intensity. We highlight that encoders for different modalities produce varying numbers of output tokens. However, our CCMT model requires an equal number of tokens for each modality. To meet the data uniformity requirement imposed by CCMT, we randomly sample a fixed number of tokens, denoted as $k \in \mathbb{N^+}$, where $k \leq k_a$. The selected tokens are then fed into the audio branch of the CCMT model. The tokens are further combined with tokens from the text modalities to facilitate the multimodal analysis. By fine-tuning the Wav2Vec2.0 model, our framework benefits from a rich representation of the audio modality, enabling the CCMT model to effectively capture and integrate both linguistic and acoustic information.

\subsection{Text branches}

To generate text transcripts from the audio samples, we utilize a set of ASR models built on top of the Whisper architecture \cite{Radford-ARXIV-2022}, comprising three distinct backbones: small, medium, and large. Employing multiple ASR models serves as a data augmentation mechanism, enhancing the training data. We underline that Whisper is a multilingual system able to generate transcriptions in several languages, depending on the language spoken in the audio files. The languages used in our speech data sets are English and French. However, our pipeline is easily applicable to other languages, if necessary. After obtaining the transcripts in the original language, we employ a language translation model called FLAN \cite{Chung-ARXIV-2022} to translate the original text into other languages. Since the translation model can introduce errors during translation, we select a group of three languages for which the translation model exhibits higher performance, namely English, French and Spanish. Although we carefully selected this group of three languages, our preliminary results (presented in the next section) show that Spanish lowers the overall performance. This influenced our decision not to use Spanish. Therefore, we incorporate only two language modalities: English and French. 

Translating a text to different languages can naturally result in a different number of words. 
Consequently, the English text given as input to the BERT model \cite{Devlin-NAACL-2019} is assumed to consist of $k_e + 1 \in \mathbb{N^+}$ tokens, comprising $k_e$ words and one class token. BERT returns the same number of tokens as output. Similarly, in the case of French, we have a total of $k_f + 1 \in \mathbb{N^+}$ tokens, comprising $k_f$ words and an extra class token. These tokens are fed into the CamemBERT model \cite{Martin-ACL-2020}, which returns $k_f+1$ output tokens. To maintain consistency across modalities, we randomly select a number of $k$ tokens from the output of the English language model (BERT) and the output of the French language model (CamemBERT), respectively. This ensures an equal number of tokens for all modalities. In the random sampling process, we make sure to always include the class tokens, since these tokens are very important for the final classification task. If the number of tokens for either modality, English or French, is below $k$, we randomly duplicate tokens until $k_e=k$ or $k_f=k$ to comply with the uniformity constraint. By incorporating both English and French language modalities, we offer the opportunity to our CCMT model to adequately capture and integrate linguistic information from multiple languages, enabling a comprehensive multimodal analysis for the target classification task.

\subsection{Cascaded Cross-Modal Transformer}

\begin{figure}[!t]
\begin{center}
\centerline{\includegraphics[width=0.7\linewidth]{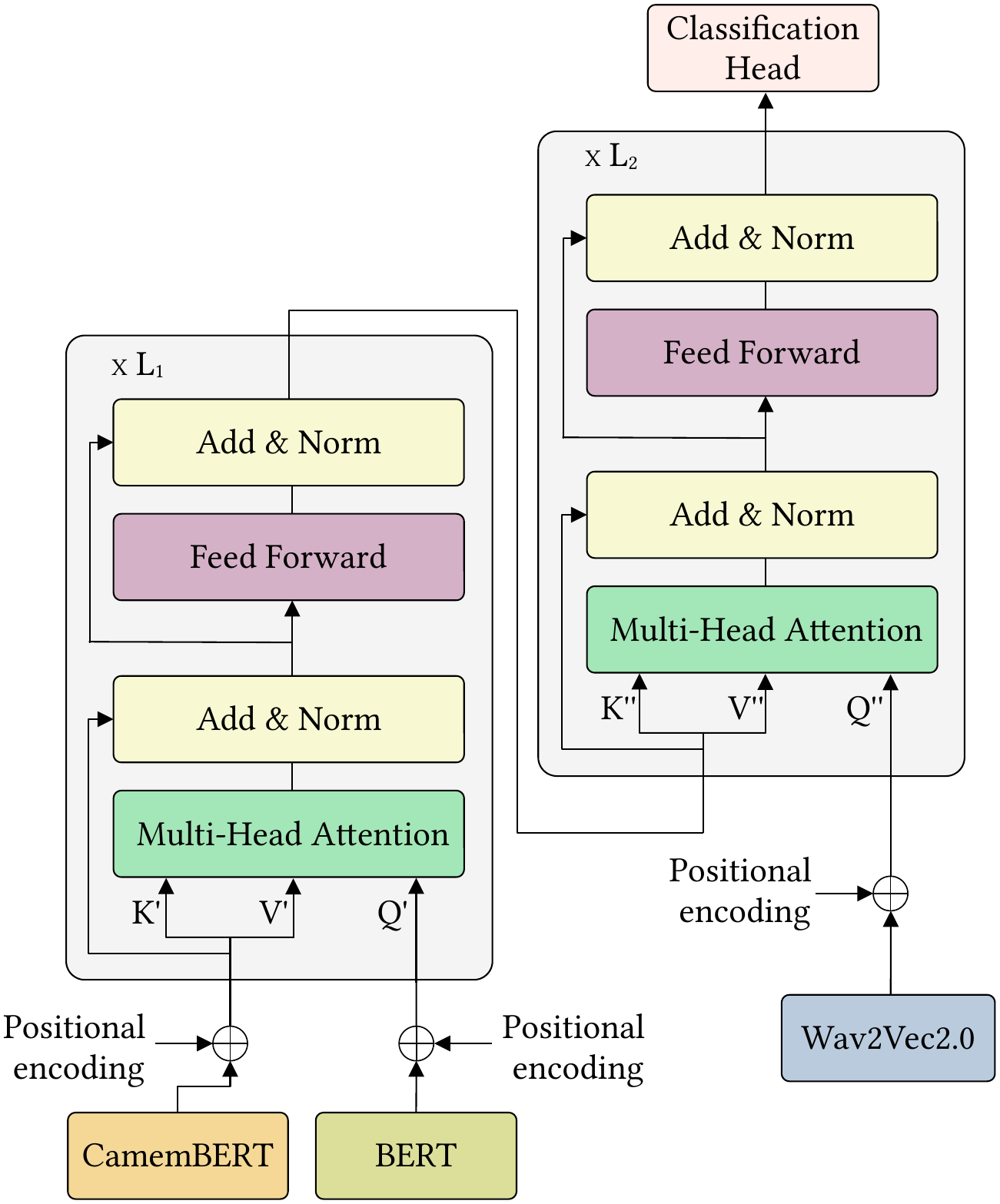}}
\vspace{0.2cm}
\caption{As input, the CCMT architecture receives tokens obtained from CamemBERT, BERT, and Wav2Vec2.0. To maintain the positional information of each modality, we introduce separate positional embeddings. The tokens are processed by two cascaded cross-attention transformer blocks. The first block combines the French and English text modalities, and the resulting tokens are combined with the audio modality. The final class token is passed to the MLP classification head to make the final predictions.}
\label{fig_ccmt}
\end{center}
\end{figure}

Considering that all models produce tokens with identical dimensionality, let $T_f \in \mathbb{R}^{k \times d}$ denote the set of tokens generated by the CamemBERT model, $T_e \in \mathbb{R}^{k \times d}$ denote the set of tokens generated by the BERT model, and $T_a \in \mathbb{R}^{k \times d}$ denote the set of audio tokens generated by the Wav2Vec2.0 model, where $d \in \mathbb{N^+}$. The audio and text tokens are combined via a cascaded cross-modal transformer, as illustrated in Figure \ref{fig_ccmt}. In order to make our model differentiate between token positions across various modalities, we introduce positional encoding vectors that are unique to each modality.

In the first transformer block, we introduce the learnable parameters $W_Q', W_K', W_V' \in \mathbb{R}^{d \times d_h}$ for the projection layers, where $d_h \in \mathbb{N^+}$ represents the dimension of a single attention head. If the original spoken language is French, we use the English modality for queries, and the French modality for keys and values, as we consider that the original language modality is more important for the final classification task. To obtain the queries, keys, and values, we perform matrix multiplications between the input tokens and the projection matrices, as follows: $Q' = T_e \cdot W_Q'$, $K' = T_f \cdot W_K'$, $V' = T_f \cdot W_V'$. If the original spoken language is English, we use the French modality for queries, and the English modality for keys and values. Consequently the query, key, and value tokens are obtained as follows: $Q' = T_f \cdot W_Q'$, $K' = T_e \cdot W_K'$, $V' = T_e \cdot W_V'$. Regardless of the original language spoken in the audio files, the output of the proposed cross-attention layer, denoted as $U' \in \mathbb{R}^{k \times d_h}$, can be expressed as follows:
\begin{equation}
\label{eq1}
U' = \mbox{softmax}\left( \frac{Q'\cdot K'^{\top}}{\sqrt{d_h}} \right) \cdot V'.
\end{equation}

Note that our cross-attention layer can have a variable number of attention heads. We further employ a learnable projection matrix $M' \in \mathbb{R}^{d_h \times d}$ to guarantee the consistency of the dimensionality of the output tokens returned by the multi-head attention layer. Multiplying the output $U'$ with $M'$ allows the output dimensionality to be restored to the original dimensionality of the input tokens. The projected output tokens are given by $Y' = U' \cdot M'$. Next, we perform the summation and normalization operations, succeeded by a feed-forward module (FF), and another summation and normalization layer. The equations formally describing these operations are presented below:
\begin{equation}
\label{eq2}
Z' = Y' + \mbox{Norm}(Y'),
\end{equation}
\begin{equation}
\label{eq3}
T_c = Z' + \mbox{FF}(\mbox{Norm}(Z')),
\end{equation}
where $T_c \in \mathbb{R}^{k \times d}$ represents the output cross-attention linguistic tokens. The first transformer block is successively repeated $L_1 \in \mathbb{N}^+$ times. In the second cross-attention transformer block, the linguistic tokens $T_c$ are combined with the audio tokens $T_a$. Here, we introduce the learnable parameters $W_Q'', W_K'', W_V'' \in \mathbb{R}^{d \times d_h}$ for the projection layers. As for the first transformer block, we obtain the queries, keys, and values by multiplying the input tokens with the corresponding projection matrices: $Q'' = T_a \cdot W_Q''$, $K''= T_c \cdot W_K''$, $V'' = T_c \cdot W_V''$. By employing operations analogous to Equations \eqref{eq1}, \eqref{eq2} and \eqref{eq3}, we obtain the output tokens $T_o \in \mathbb{R}^{k \times d}$. The second transformer block is successively repeated $L_2 \in \mathbb{N}^+$ times. The first token of the last transformer block, which represents the class token, is passed to a multi-layer perceptron (MLP) head. The MLP head predicts the final labels for the target classification task. We underline that the operations above are presented in the context of a single attention head, but the extension to multiple heads is trivial, as it immediately results from the presentation of Vaswani \etal~\cite{Vaswani-NIPS-2017}.

\section{Experiments}
\label{sec_experiments}

\subsection{Data sets}

\noindent \textbf{ComParE RSC.}
The data set supplied by the organizers of the ComParE competition \cite{Schuller-ACMMM-2023} for the Requests Sub-Challenge (RSC) is a subset of the HealthCall30 corpus, built by Lackovic \etal~\cite{Lackovic-arXiv-2022}. The data set is split into training (6,822 samples), development (3,084 samples) and test (3,503 samples). It comprises real audio recordings between call center agents and customers who called to complain regarding some issue or to request some information. Each data sample is a recording of $30$ seconds at a sampling rate of $16$ kHz. Each conversation has two separate audio channels. The first channel corresponds to the customer’s audio and the second corresponds to the agent’s audio. There are two separate binary classification tasks on this data set: request detection and complaint detection.

\noindent{\bf Speech Commands v2.}
The Speech Commands v2 (SCv2) \cite{Warden-ArXiv-2018} data set comprises spoken words, which enables the evaluation of keyword spotting systems. It consists of 105,829 audio samples, each having about one second in length. Each recording contains one of 35 common speech commands, and the task is to identify the correct command. The official split of Speech Commands v2 has 84,843 training samples, 9,981 validation samples and 11,005 test samples.

\noindent{\bf HarperValleyBank.}
The HarperValleyBank (HVB) data set \cite{Wu-ARXIV-2020} is composed of simple consumer-bank interactions, comprising roughly 23 hours of audio from 1,446 conversations between 59 different speakers. The conversations are simulated and follow realistic templates with controlled complexity and a limited vocabulary size of 700 unique words. We randomly split the data based on speaker identifiers into training (70\%), validation (15\%) and test (15\%), thus avoiding speaker leakage across subsets. We tackle two  multi-way classification tasks on HVB: action recognition (classification into 16 classes) and intent recognition (classification into 8 classes). Wu \etal~\cite{Wu-ARXIV-2020} reported the F1 score for the action recognition task, where the class distribution is highly imbalanced, and the accuracy for the intent recognition task, where the class distribution is almost uniform. Therefore, we report the same performance measures.

\subsection{Performance measures} 

In the ComParE competition, participants are evaluated and ranked using the unweighted average recall (UAR), which is the average of recall scores calculated for both positive and negative classes. Consequently, we present our performance in terms of this metric. For the Speech Commands v2 data set and the intent recognition task on the HarperValleyBank data set, we report the classification accuracy. For the action recognition task on HarperValleyBank, we report the $F_1$ score. We underline that the selection of performance measures follows the prior art \cite{Schuller-ACMMM-2023,Warden-ArXiv-2018,Ristea-INTERSPEECH-2022,Wu-ARXIV-2020} on the respective data sets.
Regardless of the evaluation measure, we repeat each experiment three times and report the average performance and the standard deviation.

\subsection{Baselines}

\noindent \textbf{ComParE RSC.}
In the audio experiments, we assess our model, built upon Wav2Vec2.0 \cite{Baevski-NeurIPS-2020}, against ResNet-50 \cite{He-CVPR-2016} and several transformer-based frameworks \cite{Gong-INTERSPEECH-2021, Ristea-INTERSPEECH-2022}. For the text experiments, we conduct a comparative analysis involving multiple NLP models \cite{Le-LREC-2020, Martin-ACL-2020} and four ASR models \cite{Baevski-NeurIPS-2020, Radford-ARXIV-2022}. Additionally, we analyze multiple aggregation methods, comparing our CCMT model with a vanilla transformer model \cite{Dosovitskiy-ICLR-2020}, as well as other common fusion methods, such as plurality (majority) voting or multi-layer perceptrons (MLPs). On the final test set, we compare with the best baseline proposed by the organizers \cite{Schuller-ACMMM-2023}, as well as the top performers in the competition \cite{Porjazovski-ACMMM-2023,Sun-ACMMM-2023}.

\noindent{\bf Speech Commands v2.}
To the best of our knowledge, there is no prior work which uses multimodal audio-textual frameworks on the Speech Commands v2 data set. Therefore, we compare our work with state-of-the-art audio models based on CNNs \cite{Majumdar-INTERSPEECH-2020} and transformers \cite{Gong-INTERSPEECH-2021, Ristea-INTERSPEECH-2022}.

\noindent{\bf HarperValleyBank.}
We include state-of-the-art results obtained with audio-only models \cite{Wu-ARXIV-2020}, text-only models \cite{Thomas-ICASSP-2022}, as well as multimodal approaches \cite{Thomas-ICASSP-2022, Sunder-ICASSP-2022}. 
    
\subsection{Data preprocessing}

\noindent \textbf{ComParE RSC.}
For ComParE RSC, we use 16 kHz audio files comprising both agent and customer speeches, which are encoded into a single channel. The single channel is obtained by averaging the original audio channels. The other preprocessing steps are identical to those of Baevski \etal~\cite{Baevski-NeurIPS-2020}. For all ASR models \cite{Baevski-NeurIPS-2020, Radford-ARXIV-2022}, we consider the audio samples that are split into two separate channels, one for the agent and one for the customer. We resample each audio channel at 16 kHz and feed it to the ASR models, following the preprocessing steps described in the original papers presenting the ASR models \cite{Baevski-NeurIPS-2020, Radford-ARXIV-2022}. For the text models, we 
simply concatenate the transcripts of the agent and the customer. Regarding the input data for the CCMT model, considering that we have a distinct number of tokens for each modality, we randomly sample $k=100$ tokens, making sure to always keep the class token for each modality.

\noindent{\bf Speech Commands v2 and HarperValleyBank.} Since the audio samples in Speech Commands v2 and HarperValleyBank are already encoded into a single channel, concatenating audio channels is not required for these two data sets. The preprocessing steps are identical to those of Baevski \etal~\cite{Baevski-NeurIPS-2020}. Similarly, for all ASR models \cite{Baevski-NeurIPS-2020, Radford-ARXIV-2022}, we follow the steps described in the original papers. We provide $k=100$ tokens per modality as input to CCMT.

\subsection{Data augmentation}
For the audio features, we employ Gaussian noise, clipping, pitch shifting, low-pass filtering, high-pass filtering and volume change as augmentation methods. For the text modalities, we use classical augmentation methods, \eg~masking and swapping words, in addition to an augmentation based on varying the ASR model. More precisely, we use randomly picked transcripts from multiple ASR models. This augmentation technique is also used for the NLP models trained on other languages, as it simply involves translating the transcripts from multiple ASR models.

\subsection{Implementation details}

We fine-tune the Wav2Vec2.0 \cite{Baevski-NeurIPS-2020} architecture on mini-batches of $16$ examples for $10$ epochs, using a learning rate of $10^{-5}$. We optimize the language models, \ie~BERT \cite{Devlin-NAACL-2019} and CamemBERT \cite{Martin-ACL-2020}, on mini-batches of $32$ examples for $25$ epochs, using a learning rate of $5\cdot10^{-5}$ and a weight decay of $10^{-5}$. For all other models, we adopt the hyperparameters recommended in the official implementations. Finally, we optimize our CCMT model on mini-batches of $32$ samples for $30$ epochs, using a learning rate of $10^{-4}$. We set the number of transformer blocks in CCMT to $L_1=L_2=8$, and the number of attention heads in each block to $8$. All models are optimized with Adam \cite{Kingma-ICLR-2014}. We motivate our design and hyperparameter choices via ablation experiments. To easily reproduce the results, we provide our code for free at: \url{https://github.com/ristea/ccmt}.

\begin{table*}[!t]
  \caption{Results on the ComParE RSC development set, featuring various audio-based models. The models are either trained from scratch or fine-tuned. The architectures marked with $*$ use pretrained weights. We present the mean UAR (in percentages) and the standard deviation over three runs, using bold font to indicate the best score on each task.}
  \label{tab_audio}
  \begin{center}
  \begin{tabular}{llcc}
    \toprule
    \multirow{2}{*}{Model}       & \multirow{2}{*}{Input data}    & \multicolumn{2}{c}{UAR} \\
    \cmidrule{3-4}
    & & Request & Complaint\\
    \midrule
    ResNet-50 \cite{He-CVPR-2016}         & Spectrogram      &  $59.51 \pm 1.27$ & $52.18 \pm 0.86$ \\
    ResNet-50 \cite{He-CVPR-2016}         & STFT      &  $60.84 \pm 1.08$   & $53.49 \pm 0.73$ \\
    ResNet-50 \cite{He-CVPR-2016}         & Mel-Spectrogram    &  $60.31 \pm 1.01$  & $53.44 \pm 0.74$ \\
    SepTr \cite{Ristea-INTERSPEECH-2022}      & STFT     &  $62.31 \pm 0.59$   & $54.03 \pm 0.55$ \\
    AST$^*$ \cite{Gong-INTERSPEECH-2021}    & Spectrogram    &  $64.72 \pm 0.45$   & $55.91 \pm 0.39$ \\
    1D Transformer & Time domain    &  $ 61.63 \pm 0.42$   &  $53.82 \pm 0.39$ \\
    Wav2Vec2.0 \cite{Baevski-NeurIPS-2020} & Time domain   &  $ 68.87 \pm 0.21$   & $ 56.55 \pm 0.22$ \\
    Wav2Vec2.0$^*$ \cite{Baevski-NeurIPS-2020} & Time domain    & $ \textbf{71.64} \pm 0.16$   & $ \textbf{58.12} \pm 0.16$ \\
    \bottomrule
  \end{tabular}
  \end{center}
\end{table*}

\subsection{Results on the ComParE Requests Sub-Challenge}

\subsubsection{Preliminary results for the audio modality}

In Table \ref{tab_audio}, we outline the results of models utilizing the audio modality. Among the assessed architectures, the ResNet-50 \cite{He-CVPR-2016} model exhibits subpar performance, irrespective of the given input representation. In contrast, transformer-based models consistently reach higher performance. Notably, the pre-trained AST \cite{Gong-INTERSPEECH-2021} model attains a Request UAR of $64.72\%$ and a Complaint UAR of $55.91\%$, outperforming the alternative transformer architectures, namely SepTr \cite{Ristea-INTERSPEECH-2022} and the 1D transformer. Nevertheless, the most optimal outcomes are achieved by the Wav2Vec2.0 \cite{Baevski-NeurIPS-2020} model. Indeed, we obtain a Request UAR of $71.64\%$ and a Complaint UAR of $58.12\%$, just by fine-tuning the Wav2Vec2.0 model. This highlights the effectiveness of the audio representation based on the temporal domain used by Wav2Vec2.0. These findings underscore the advantages of employing pretrained models in audio classification tasks. Based on the results presented in Table \ref{tab_audio}, we opt for the fine-tuned Wav2Vec2.0 model for integration into our multimodal pipeline.


\begin{table*}[!t]
  \caption{Results on the ComParE RSC development set using various language models fine-tuned on French transcripts generated by Wav2Vec2.0 \cite{Baevski-NeurIPS-2020} and Whisper \cite{Radford-ARXIV-2022} ASR models. Whisper S+M+L denotes our augmentation technique, incorporating transcripts from all three ASR models (small, medium and large). The Wav2Vec2.0 and Whisper models are frozen. We present the mean UAR (in percentages) and the standard deviation over three runs, using bold font to indicate the best score on each task.}
  \label{tab_text}
  \begin{center}
  \begin{tabular}{llcc}
    \toprule
    \multirow{2}{*}{Model}       & \multirow{2}{*}{ASR model}    & \multicolumn{2}{c}{UAR} \\
    \cmidrule{3-4}
    & & Request & Complaint\\
    
    \midrule
    LSTM  \cite{Hochreiter-NC-1997}               & Wav2Vec2.0 \cite{Baevski-NeurIPS-2020}       &    $71.14 \pm 0.51$   & $55.49 \pm 0.50$ \\
    FlauBERT \cite{Le-LREC-2020}           & Wav2Vec2.0 \cite{Baevski-NeurIPS-2020}       &    $76.82 \pm 0.21$   & $58.77 \pm 0.23$ \\
    CamemBERT \cite{Martin-ACL-2020}          & Wav2Vec2.0 \cite{Baevski-NeurIPS-2020}       &    $77.45 \pm 0.13$   & $60.15 \pm 0.11$ \\
    \midrule
    CamemBERT \cite{Martin-ACL-2020}          & Whisper S \cite{Radford-ARXIV-2022}        &    $79.71 \pm 0.19$   & $62.92 \pm 0.20$ \\
    CamemBERT  \cite{Martin-ACL-2020}         & Whisper M \cite{Radford-ARXIV-2022}        &    $81.86 \pm 0.11$   & $64.83 \pm 0.11$ \\
    CamemBERT \cite{Martin-ACL-2020}          & Whisper L  \cite{Radford-ARXIV-2022}        &    $82.03 \pm 0.10$   & $65.47 \pm 0.09$ \\
    CamemBERT  \cite{Martin-ACL-2020}         & Whisper S+M+L     &    $\textbf{82.44} \pm 0.08$   & $\textbf{65.61} \pm 0.08$ \\
    \bottomrule
  \end{tabular}
  \end{center}
\end{table*}

\subsubsection{Preliminary results for the French text modality}

We outline the results obtained for the French text transcripts with various NLP models in Table~\ref{tab_text}. We explore various ASR models, including Wav2Vec2.0 (large)\footnote{\url{https://huggingface.co/facebook/wav2vec2-large-xlsr-53-french}} \cite{Baevski-NeurIPS-2020} and three versions of the Whisper architecture \cite{Radford-ARXIV-2022} (small, medium, and large). When using the Wav2Vec2.0 ASR model, the CamemBERT model \cite{Martin-ACL-2020} achieves the best performance, with a Request UAR of $77.45\%$ and a Complaint UAR of $60.15\%$. For the subsequent experiments, we thus select the CamemBERT model in detriment of the LSTM and FlauBERT language models.

Switching to the family of Whisper ASR models results in considerable performance gains. Actually, we attain the best performance when the transcripts generated by all Whisper models are jointly used. The corresponding performance levels are a request UAR of $82.44\%$ and a complaint UAR of $65.61\%$.  We underline that the performance levels reported for the text modality in Table \ref{tab_text} are significantly higher than the audio modality results presented in Table \ref{tab_audio}. This indicates that, for request and complaint detection, language features play a more significant role in reaching high performance than acoustic features.

\begin{table*}[!t]
  \caption{Results on the ComParE RSC development set, featuring diverse language models operating across three different languages: English (En), French (Fr), and Spanish (Sp). We present the results obtained by aggregating CamemBERT with other language models through an MLP-based aggregation method. The reported metrics are the mean UAR (in percentages) and the standard deviation over three runs. We use bold font to indicate the best score on each task.}
  \label{tab_language}
  \begin{center}
  \begin{tabular}{lccc}
    \toprule
    \multirow{2}{*}{Model}       & \multirow{2}{*}{Language}    & \multicolumn{2}{c}{UAR} \\
    \cmidrule{3-4}
    & & Request & Complaint\\
    \midrule
    CamemBERT \cite{Martin-ACL-2020}   & Fr  &  $82.44 \pm 0.08$   & $65.61 \pm 0.08$  \\
    RoBERTa \cite{Liu-ARXIV-2019}            & En    &    $78.57 \pm 0.07$   & $63.89 \pm 0.10$ \\
    BERT \cite{Devlin-NAACL-2019}              & En    &    $79.35 \pm 0.08$   & $63.91 \pm 0.08$ \\
    BERT \cite{Devlin-NAACL-2019}              & Sp    &    $72.41 \pm 0.08$   & $59.87 \pm 0.11$ \\
    \midrule
    CamemBERT \cite{Martin-ACL-2020}+BERT \cite{Devlin-NAACL-2019}   & Fr+En  &  $\mathbf{82.61} \pm 0.08$   & $\mathbf{65.91} \pm 0.08$ \\
    CamemBERT \cite{Martin-ACL-2020}+BERT \cite{Devlin-NAACL-2019}   & Fr+Sp  &  $81.80 \pm 0.08$   & $64.11 \pm 0.09$ \\
    CamemBERT \cite{Martin-ACL-2020}+2$\times$BERT \cite{Devlin-NAACL-2019}  & Fr+En+Sp  &  $82.01 \pm 0.08$   & $64.95 \pm 0.09$ \\
    \bottomrule
  \end{tabular}
  \end{center}
\end{table*}

\subsubsection{Preliminary results for multiple text modalities}

In Table~\ref{tab_language}, we showcase the results of language transformers across three distinct languages: French, English, and Spanish. Among the examined models, the CamemBERT \cite{Martin-ACL-2020} language model, which is trained on French transcripts, attains the highest performance. This is an anticipated result, given that the audio calls are in French, and translating them to new languages may introduce errors and compromise performance. Nonetheless, we hypothesize that combining models pretrained on a diverse set of languages can enhance the effectiveness of the CamemBERT model. Consequently, we explore multiple combinations of the CamemBERT model with other models trained on English and Spanish, fusing the distinct language models via an MLP block. For the English and Spanish languages, we fine-tune BERT \cite{Devlin-NAACL-2019} models on transcripts translated to the respective languages. The fine-tuned models start from weights pretrained on English and Spanish data, respectively. Interestingly, the BERT model trained on English data surpasses its Spanish counterpart by roughly $6\%$ and $4\%$ in terms of UAR for the request and complaint classes, respectively. In the fusion experiments, the most favorable results are achieved by combining the French and English models, surpassing the results of the CamemBERT model by around $0.3\%$ in terms of UAR for both request and complaint classes. Nevertheless, the inclusion of the Spanish BERT model results in a performance decline for both classes. Consequently, we omit the Spanish language model from the following experiments.

\begin{table*}[!t]
  \caption{Results on the ComParE RSC development set, featuring various methods to aggregate three models: a Wav2Vec2.0 model fine-tuned on audio samples, and two language models fine-tuned on French [Fr] and English [En] text samples, respectively. We report the mean UAR (in percentages) and the standard deviation over three runs, using bold font to indicate the best score on each task.}
  \label{tab_fusion}
  \begin{center}
  \begin{tabular}{lccccc}
    \toprule
    \multirow{2}{*}{Model}       & \multirow{2}{*}{Text [Fr]}  & \multirow{2}{*}{Text [En]}  & \multirow{2}{*}{Audio} & \multicolumn{2}{c}{UAR} \\
    \cmidrule{5-6}
    & & & & Request & Complaint\\
    \midrule
    Majority voting      & $\checkmark$ & $\checkmark$ & $\checkmark$  &    $80.08 \pm 0.11$   & $62.11 \pm 0.13$ \\
    \midrule
    MLP          & $\checkmark$ &  & $\checkmark$    &    $82.60 \pm 0.07$   & $65.98 \pm 0.07$ \\
    MLP          & $\checkmark$ & $\checkmark$ &     &    $82.61 \pm 0.08$   & $65.91 \pm 0.08$ \\
    MLP          & $\checkmark$ & $\checkmark$ & $\checkmark$    &    $82.65 \pm 0.08$   & $66.08 \pm 0.07$ \\
\midrule
    Transformer           & $\checkmark$ &  & $\checkmark$  &    $82.81 \pm 0.08$   & $65.99 \pm 0.09$ \\
    Transformer           & $\checkmark$ &  $\checkmark$  &  &   $82.04 \pm 0.07$   & $65.24 \pm 0.08$ \\
    Transformer           & $\checkmark$ & $\checkmark$ & $\checkmark$  &    $82.81 \pm 0.09$   & $66.13 \pm 0.07$ \\

    \midrule
    CCMT (ours)    & $\checkmark$ &    & $\checkmark$ &   $83.01 \pm 0.08$   & $66.20 \pm 0.07$ \\
    CCMT (ours)    & $\checkmark$ &  $\checkmark$  &  &   $81.96 \pm 0.08$   & $65.84 \pm 0.09$ \\
    CCMT (ours)    & $\checkmark$ &  $\checkmark$  & $\checkmark$ &   $\textbf{83.31} \pm 0.08$   & $\textbf{66.64} \pm 0.08$ \\
    \bottomrule
  \end{tabular}
  \end{center}
\end{table*}

\subsubsection{Comparative results for multimodal methods}

In Table \ref{tab_fusion}, we showcase the outcome of the multimodal experiments based on combining three models: Wav2Vec2.0 trained on audio data, CamemBERT \cite{Martin-ACL-2020} trained on French  transcripts, and BERT \cite{Devlin-NAACL-2019} trained on English translations. Upon combining the different modalities, the following notable patterns emerge, regardless of the chosen fusion approach. Combining CamemBERT with Wav2Vec2.0 \cite{Baevski-NeurIPS-2020} proves to be more effective than fusing the CamemBERT and BERT language models. The best performance levels are consistently obtained when we fuse all three modalities together. Concerning the results of the alternative aggregation techniques, we observe that conventional approaches, namely majority voting and MLP, reach lower performance levels than more intricate methods based on transformers. Fusing tokens from all modalities via a transformer model leads to an UAR of $82.81\%$ for the request class and $66.13\%$ for the complaint class, respectively. Nonetheless, we reach the highest performance levels by combining all modalities via the proposed CCMT model. These results confirm that CCMT represents a powerful technique to combine audio-textual representations.

\begin{figure*}[!t]
\begin{center}
\centerline{\includegraphics[width=0.98\linewidth]{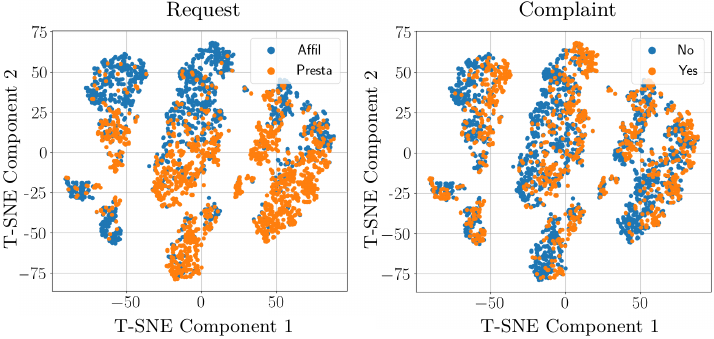}}
\vspace{0.2cm}
\caption{A t-SNE visualization of the CCMT embedding space for the ComParE RSC development set. On the left-hand side, the data points are labeled according to the request classes. On the right-hand side, the labels represent the complaint classes. Best viewed in color.}
\label{fig_tsne}
\end{center}
\end{figure*}

\begin{table*}[!t]
  \caption{Results on the ComParE RSC private test set obtained by our CCMT model with two or three input modalities. There are three state-of-the-art competitors, one proposed by the ComParE RSC organizers \cite{Schuller-ACMMM-2023} and two proposed by other RSC participants \cite{Sun-ACMMM-2023,Porjazovski-ACMMM-2023}. Based on the results reported on the ComParE RSC private test, our method (CCMT) was declared the winner of the Requests Sub-Challenge of ComParE 2023 (see \url{http://www.compare.openaudio.eu/winners/}). We report the UAR in percentages, using bold font to indicate the best score on each task.}
  \label{tab_private}
  \begin{center}
  \setlength\tabcolsep{3.3pt}
  \begin{tabular}{lcccc}
    \toprule
    \multirow{2}{*}{Method}       & \multirow{2}{*}{Modalities}    & \multicolumn{3}{c}{UAR} \\
    \cmidrule{3-5}
     & & Request & Complaint & Overall \\
    \midrule
    Schuller \etal~\cite{Schuller-ACMMM-2023} & Audio & $67.2$   & $52.9$ & $60.1$ \\  
    Sun \etal~\cite{Sun-ACMMM-2023} & Audio & $63.4$ & $55.4$ & $59.4$ \\  
    Porjazovski \etal~\cite{Porjazovski-ACMMM-2023}  & Text [Fr]+Audio & $85.4$ & $60.2$ & $72.8$ \\ 
    \midrule
    CCMT (ours)  & Text [Fr]+Audio  &  $85.09$   & $64.73$ & $74.91$ \\
    CCMT (ours)  & Text [Fr]+Text [En]+Audio  &  $\textbf{85.87}$   & $\textbf{65.41}$ & $\textbf{75.64}$ \\
    \bottomrule
  \end{tabular}
  \end{center}
\end{table*}

To further assess the discriminative power of CCMT, we illustrate a t-SNE visualization of the feature space derived from the class tokens of our top-scoring CCMT model in Figure \ref{fig_tsne}. The visualization reveals the presence of seven distinct inner clusters, with each cluster containing samples from both classes for both classification tasks, complaint and request. Notably, there is a clear (almost linear) separation within each cluster between samples belonging to distinct request classes. At the same time, it looks like distinguishing samples for the complaint detection task is more challenging due to the less evident separation of the complaint classes. This is consistent with the results reported so far on the two tasks, \ie~requests are generally easier to detect than complaints.

\begin{table*}[!t]
  \caption{Ablation study on the ComParE RSC development set, considering different configurations of the CCMT architecture trained on all three modalities. We vary the dimension of the projection layer, the number of transformer blocks (depth), the number of cross-attention heads, the token dimension in each head, and the token sampling strategy. We report the mean UAR (in percentages) and the standard deviation over three runs, using bold font to indicate the best score on each task.}
  \label{tab_ablation}
  \begin{center}
  \setlength\tabcolsep{5.5pt}
  \begin{tabular}{ccccccc}
    \toprule
    Projection  & \multirow{2}{*}{Depth} & \multirow{2}{*}{Heads} &   Head & Random & \multicolumn{2}{c}{UAR} \\
    \cmidrule{6-7}
    dimension & & & dimension & selection & Request & Complaint\\
    
    \midrule
    256    & 6 &  8  & 512 &  $\checkmark$ & $ 81.14 \pm 0.11$   & $62.98 \pm 0.13$ \\
    512    & 6 &  8  & 512 &   $\checkmark$ &$ 81.45 \pm 0.10$   & $63.21 \pm 0.11$ \\
    -      & 6 &  8  & 512 &   $\checkmark$ &$ 82.97 \pm 0.08$   & $66.18 \pm 0.09$ \\
\midrule
    -    & 7 &  8  & 512 &  $\checkmark$ & $ 83.28 \pm 0.08$   & $66.11 \pm 0.07$ \\
    -    & 8 &  8  & 512 &  $\checkmark$ & $ \textbf{83.31} \pm 0.08$   & $\textbf{66.64} \pm 0.08$ \\
    -    & 9 &  8  & 512 &  $\checkmark$ & $ 83.22 \pm 0.08$   & $65.99 \pm 0.08$ \\
\midrule
    -    & 8 &  7  & 512 & $\checkmark$ &  $ 83.11 \pm 0.07$   & $66.63 \pm 0.08$ \\
    -    & 8 &  9  & 512 &  $\checkmark$ & $ 83.19 \pm 0.08$   & $66.49 \pm 0.07$ \\
\midrule
    -    & 8 &  8  & 256 &  $\checkmark$ & $ 83.22 \pm 0.08$   & $66.02 \pm 0.08$ \\
    -    & 8 &  8  & 1024 & $\checkmark$ & $ 83.01 \pm 0.08$   & $65.31 \pm 0.09$ \\
\midrule
    -    & 8 &  8  & 512 &  & $82.51 \pm 0.11$   & $66.13 \pm 0.09$ \\

\bottomrule
  \end{tabular}
  \end{center}

\end{table*}

\subsubsection{Results on the ComParE RSC private test set}

We present the performance levels on the ComParE RSC private test set in Table \ref{tab_private}. We submitted two models for the final evaluation. Our first submission is based on a CCMT model that aggregates two data modalities, represented by CamemBERT and Wav2Vec2.0 tokens. Our second submission is produced by our full CCMT model, which comprises three modalities. 
When using only two data modalities (French transcripts + audio files), the performance is visibly lower than the performance obtained by the full CCMT model. This observation confirms the beneficial impact of integrating models trained on distinct languages.  
Our best submission reaches an average UAR of $75.64\%$, surpassing the second-best model, which is proposed by Porjazovski \etal~\cite{Porjazovski-ACMMM-2023}, by $2.84\%$. The third-best model \cite{Sun-ACMMM-2023} is only based on the provided audio modality, attaining even lower results. The same can be said about the baseline proposed by the organizers \cite{Schuller-ACMMM-2023}. In summary, the results presented in Table \ref{tab_private} clearly confirm that the idea of generating multiple text modalities via ASR and NMT is beneficial. The final standings also show the superiority of our cascaded cross-modal transformer over competing approaches.

\subsubsection{Ablation study}

To analyze the impact of different configurations on the performance of our model, we conduct an ablation study and present the corresponding results in Table \ref{tab_ablation}.
One important observation is that our model deviates from a regular transformer architecture by omitting the input projection layer. The results show that including a learnable projection over the input tokens has a considerable negative effect on the performance. This suggests that the language and audio models extract meaningful information, alleviating the need of having another projection layer over the tokens.

Furthermore, we explore the impact of varying the depth, the number of attention heads, and the token dimension in each head of the CCMT architecture. Interestingly, we find that these variations do not lead to significant performance changes. The best results are obtained with a depth of 8 blocks ($L_1=L_2=8$), each with 8 attention heads and a dimension of 512 for the tokens. 

In the proposed CCMT architecture, after the multi-head attention module, we sum the result with the initial tokens, which generated the keys and values. Considering that the output of the multi-head attention layer has the same length as the query, we cannot sum it with the initial tokens (which generated the keys and values), without having uniformity across the length dimension. In order to perform the ablation study without the random selection strategy, we need to modify the CCMT architecture to be able to handle distinct token dimensions across all data modalities. More exactly, we change the summation operation after both multi-head attention layers inside CCMT, such that the summation is performed using the initial query tokens, instead of the corresponding key and value tokens. We observe that the results are slightly worse, highlighting the efficiency of our architecture based on random token selection.

Overall, the ablation study emphasizes the importance of removing the input projection in the CCMT model, while indicating that other variations have a relatively minor performance impact.


\begin{table*}[!t]
  \caption{Accuracy rates (in percentages) of our framework versus various state-of-the-art methods on the Speech Commands v2 data set. We report results with CCMT based on an audio model and two language models trained on English [En] and French [Fr] transcripts. The best score is highlighted in bold.}
  \label{tab_scv2}
  \begin{center}
  \begin{tabular}{lcc}
    \toprule
    Method & Modalities    & Accuracy \\
    \midrule
    MatchboxNet \cite{Majumdar-INTERSPEECH-2020} & Audio & $97.40$ \\
    AST \cite{Gong-INTERSPEECH-2021} & Audio & $98.11 \pm 0.05$ \\
    SepTr \cite{Ristea-INTERSPEECH-2022} & Audio & $98.51 \pm 0.07$ \\
    \midrule
    Wav2Vec2.0 & Audio  &  $98.32 \pm 0.05$   \\
    BERT & Text [En]  &  $95.44 \pm 0.06$ \\
    CamemBERT & Text [Fr]  &  $95.30 \pm 0.08$ \\
    CCMT (ours) & Text [En]+Text [Fr] &  $95.43 \pm 0.06$   \\
    CCMT (ours) & Audio+Text [En] &  $98.84 \pm 0.05$   \\
    CCMT (ours) & Audio+Text [En]+Text [Fr]  &  $\textbf{98.86} \pm 0.04$   \\

    \bottomrule
  \end{tabular}
  \end{center}
\end{table*}

\subsection{Results on the Speech Commands v2 data set}

In Table \ref{tab_scv2}, we present the results of the proposed framework versus the results of several state-of-the-art methods on the Speech Commands v2 data set. We include results for multiple modalities, such as audio-only, text-only and distinct combinations of multimodal inputs. To the best of our knowledge, there is no previous work that uses audio-textual data on this data set. Therefore, we can only compare with audio-based state-of-the-art methods proposed in recent studies \cite{Majumdar-INTERSPEECH-2020, Gong-INTERSPEECH-2021, Ristea-INTERSPEECH-2022}. We observe that the Wav2Vec2.0 model surpasses two recent methods \cite{Majumdar-INTERSPEECH-2020, Gong-INTERSPEECH-2021}, while the text-only BERT achieves lower results. However, our audio-textual model outperforms all state-of-the-art methods, regardless of the inclusion of the French text modality. When we combine the English and French text modalities, there is no statistically significant gain over using the English modality alone. The main reason behind this fact is that the data set mostly consists of single-word samples, and changing the language by translation brings no significant contribution. Moreover, the ASR models fail in edge case scenarios, \eg~trimmed or muffled words, thus affecting the overall accuracy. To verify this statement, we compute the accuracy by detecting if the words are present in the transcript. The resulting accuracy rate is $94.98\%$.
Our best accuracy of $98.86\%$ is obtained by combining audio and text data in both English and French. With this score, CCMT sets a new accuracy record on the Speech Commands v2 data set.

\begin{table*}[!t]
  \caption{Results of our framework versus various state-of-the-art methods on the HarperValleyBank data set. We report results with CCMT based on an audio model and two language models trained on English [En] and French [Fr] transcripts. We report the $F_1$ score for the action recognition task, and the classification accuracy (in percentages) for the intent recognition task, using bold font to indicate the best score on each task.}
  \label{tab_harper}
  \begin{center}
  \setlength\tabcolsep{4.8pt}
  \begin{tabular}{lccc}
    \toprule
    \multirow{2}{*}{Method} & \multirow{2}{*}{Modalities}    & Action & Intent \\  
    
     &     & ($F_1$) & (Accuracy)\\

    \midrule
    CTC \cite{Wu-ARXIV-2020} & Audio & $0.3864$ & $45.47$ \\
    LAS \cite{Wu-ARXIV-2020} & Audio & $0.2931$ & $34.96$ \\
    MLT \cite{Wu-ARXIV-2020} & Audio & $0.3222$ & $42.28$ \\
    ASR RNN-T \cite{Thomas-ICASSP-2022} & Text & $0.5357$ & - \\
    ASR RNN-T \cite{Thomas-ICASSP-2022} & Audio+Text & $0.5495$ & - \\
    HIER-ST \cite{Sunder-ICASSP-2022} & Audio+Text & $0.6170$ & - \\
    \midrule
    Wav2Vec2.0 & Audio  &  $0.3745 \pm 0.0007$ & $46.10 \pm 0.79$   \\
    BERT & Text [En]  &  $0.5396 \pm 0.0008$ & $33.67 \pm 1.21$  \\
    CamemBERT & Text [Fr]  &  $0.5281 \pm 0.0007$ & $32.92 \pm 0.98$  \\
    CCMT (ours) & Text [En]+Text [Fr] &  $0.5466 \pm 0.0006$ & $36.45 \pm 1.18$    \\
    CCMT (ours) & Audio+Text [En] &  $0.6098 \pm 0.0006$ & $46.02 \pm 0.65$   \\
    CCMT (ours) & Audio+Text [En]+Text [Fr]  &  $\textbf{0.6246} \pm 0.0007$ & $\textbf{47.56} \pm 0.74$    \\

    \bottomrule
  \end{tabular}
  \end{center}
\end{table*}

\subsection{Results on the HarperValleyBank data set}

In Table \ref{tab_harper}, we present the results of the proposed framework on the HarperValleyBank data set. We include results for both independent and combined modalities. We compare with both audio-based \cite{Wu-ARXIV-2020} and audio-textual \cite{Thomas-ICASSP-2022, Sunder-ICASSP-2022} methods from recent literature. We generally observe that audio-only frameworks \cite{Wu-ARXIV-2020}, including Wav2Vec2.0, obtain lower performance levels with respect to multimodal systems. For example, the ASR RNN-T model \cite{Thomas-ICASSP-2022} reaches a better $F_1$ score with the audio-textual version than with its text-only version. Regarding our models based on CCMT, we note that adding the French text modality is generally helpful for both text-only and audio-textual versions. This further confirms our conjecture stating that, for complex audio classification tasks, generating textual information in multiple languages can significantly improve the overall performance. Our best CCMT model is based on mixing audio, English text, and French text. This combination attains an $F_1$ score of $0.6246$ for action recognition and an accuracy of $47.56\%$ for intent recognition, surpassing all state-of-the-art methods included in Table \ref{tab_harper}. 

\section{Conclusion}

In this study, we proposed CCMT, a novel multimodal transformer-based architecture designed for speech classification tasks. Our framework is based on automatically generating text representations via pretrained ASR and NMT models, transforming the original speech classification task into an audio-textual (multimodal) classification task. CCMT integrates an audio model and two distinct language models, effectively aggregating information from different modalities. This is achieved through cascaded cross-attention transformer blocks, which iteratively combine information from linguistic and audio features. We evaluated the performance of CCMT on three distinct data sets, namely Speech Commands v2, HarperValleyBank and the Requests Sub-Challenge of the ACM Multimedia 2023 Computational Paralinguistics Challenge \cite{Schuller-ACMMM-2023}. Our framework demonstrated outstanding results on all data sets. Remarkably, CCMT was ranked on the first place in the ComParE challenge with an average UAR of $75.64\%$. CCMT significantly outperformed the competition baseline (by more than $15\%$) and other competitors (by more than $2.84\%$), indicating the effectiveness of our approach. The success of CCMT can be attributed to its ability to harness the complementary nature of linguistic and acoustic features. By combining information from multiple audio-textual modalities, our framework captures a more robust representation of the input data, leading to enhanced performance.

\backmatter

\section*{Declarations}

\textbf{Funding} The authors did not receive support from any organization for the submitted work.

\vspace{0.5cm}
\noindent
\textbf{Conflict of interest} The authors have no conflicts of interest to declare that are relevant to the content of this article.

\vspace{0.5cm}
\noindent
\textbf{Ethics approval} Not applicable.

\vspace{0.5cm}
\noindent
\textbf{Consent to participate} Not applicable.

\vspace{0.5cm}
\noindent
\textbf{Consent for publication} The authors give their consent for publication.

\vspace{0.5cm}
\noindent
\textbf{Authors' contributions} Nicolae-C\u{a}t\u{a}lin Ristea implemented the method, performed the experiments, and wrote the preliminary draft of the paper. Radu Tudor Ionescu designed the method and the experiments. Radu Tudor Ionescu and Andrei Anghel revised the manuscript.

\vspace{0.5cm}
\noindent
\textbf{Availability of data and materials} The datasets are publicly available online.

\vspace{0.5cm}
\noindent
\textbf{Code availability} The code has been made publicly available for non-commercial use at \url{https://github.com/ristea/ccmt}.

\vspace{0.5cm}
\noindent
\textbf{Open Access} This article is licensed under a Creative Commons Attribution 4.0 International License, which permits use, sharing, adaptation, distribution and reproduction in any medium or format, as long as you give appropriate credit to the original author(s) and the source, provide a link to the Creative Commons license, and indicate if changes were made. The images or other third party material in this article are included in the article’s Creative Commons license, unless indicated otherwise in a credit line to the material. If material is not included in the article’s Creative Commons license and your intended use is not permitted by statutory regulation or exceeds the permitted use, you will need to obtain permission directly from the copyright holder. To view a copy of this license, visit \url{http://creativecommons.org/licenses/by/4.0/}.





\bibliography{sn-bibliography}

\end{document}